\documentclass[letterpaper, 10 pt, conference]{ieeeconf}  

\IEEEoverridecommandlockouts                              

\IEEEoverridecommandlockouts

\usepackage{cite}
\usepackage{amsmath,amssymb,amsfonts}
\usepackage{algorithmic}
\usepackage{graphicx}
\usepackage{textcomp}
\usepackage{subcaption}
\usepackage{xcolor}
\usepackage{hyperref}
\usepackage{booktabs}
\hypersetup{
    colorlinks=true,
    linkcolor=black,
    urlcolor=blue,
    citecolor=black,
    pdfborder={0 0 0}
}

\title{\LARGE \bf
A New Implementation of NeoSLAM and a Comparative Evaluation with RatSLAM
}

\author{Joao Victor T. Borges$^{1}$, Fabio Coelho$^{1}$, Paulo Padrao$^{2}$, Jose Fuentes$^{3}$, Ramon R. Costa$^{1}$, Liu Hsu$^{1}$ \\ and Leonardo Bobadilla$^{3}$
\thanks{Fabio Coelho, Joao Victor T. Borges, Ramon Romankevicius, and Liu Hsu are with the Department of Electrical Engineering, Federal University of Rio de Janeiro, Brazil
        {\tt\footnotesize f.coelho@gsuite.iff.edu.br, \{borgesjvt, ramonrcosta\}@gmail.com, lhsu@coppe.ufrj.br}}%
\thanks{$^{2}$ Paulo Padrao is with the Department of Mathematics and Computer Science, Providence College, Providence, RI, USA
{\tt\footnotesize ppadraol@providence.edu}}%
\thanks{$^{3}$ Jose Fuentes and Leonardo Bobadilla are with the School of Computing and Information Sciences, Florida International University, Miami, FL, USA
{\tt\footnotesize jfuen099@fiu.edu, bobadilla@cs.fiu.edu}}
}

\begin{document}

\maketitle
\thispagestyle{empty}
\pagestyle{empty}


\begin{abstract}
This paper presents a new implementation of the NeoSLAM algorithm. The proposed version is a complete rewrite of NeoSLAM into a modular architecture using modern frameworks that, together, enable real-time execution with minimal discarding of input data. This work also provides a comparative evaluation between NeoSLAM and RatSLAM across three datasets under varying environmental conditions. The experimental results highlight differences in mapping consistency and trajectory reconstruction, demonstrating the effectiveness and practical applicability of the proposed ROS 2-based implementation. The results indicate that the new NeoSLAM outperforms the original in terms of processing throughput for real-time applications and achieves comparable performance to RatSLAM in terms of map reconstruction across the evaluated datasets.
\end{abstract}


\section{Introduction}

Autonomous mobile robotics faces substantial challenges in real-world applications, particularly when operating in unknown, dynamic, or GPS-denied environments. Such scenarios, ranging from subterranean exploration to forest environments and underwater navigation, demand reliable and accurate localization and mapping capabilities \cite{baxevani2022}. In these scenarios, robots must operate without prior maps while coping with perceptual aliasing, sensor noise, and environmental changes. Reliable localization and mapping are therefore critical to ensure safe and efficient operation, especially in long-term deployments where uncertainty accumulates over time \cite{thrun2006}.

Simultaneous Localization and Mapping (SLAM) addresses this challenge by enabling a robot to build a map while estimating its own pose within that map. By fusing sensory observations with motion models, SLAM algorithms iteratively reduce state uncertainty and maintain global consistency. Loop closure detection further mitigates accumulated drift and plays a key role in long-term autonomy and large-scale exploration \cite{thrun2006,orbslam2015}.

Among the various SLAM paradigms, biologically inspired approaches have gained attention, beginning with RatSLAM, which draws inspiration from the rodent hippocampal navigation system \cite{ratslam2004, ratslam2008}. Subsequent models incorporated findings related to place cells, grid cells, and head direction cells to improve spatial representation and robustness \cite{hippocampus1978, headcells1984, entorhinal2005, hippocampus2006}. Compared to classical probabilistic SLAM methods, these approaches often provide scalable topological representations and improved tolerance to perceptual ambiguity in visually complex environments, allowing them to operate in scenarios where traditional probabilistic methods often fail completely \cite{ratslam2008}.

Building upon these foundations, NeoSLAM was recently proposed as a neuroscience-based model motivated by advances in understanding the neocortex and hippocampal–entorhinal circuits \cite{neoslam2024}. Unlike systems primarily inspired by hippocampal mechanisms and typically modeled using a Continuous Attractor Neural Network (CANN), NeoSLAM also integrates neocortical principles, including a sequence memory model based on Hierarchical Temporal Memory (HTM) \cite{htm2016} and Sparse Distributed Representations (SDRs) \cite{sdr2015}. This hybrid design enables improved sequence learning, adaptability to appearance changes, and enhanced long-term visual place recognition performance under varying environmental conditions.

The original NeoSLAM $^{\footnotemark}$\footnotetext{\url{https://github.com/cappizzino/neoslam_ws}} implementation was developed within the Robot Operating System (ROS) 1 workspace. Although functional, this implementation presents architectural limitations that create bottlenecks in real-time data processing, preventing the system from handling all incoming sensory data at full rate. Furthermore, the modeling of SDRs and HTM relied on the \texttt{NuPIC} $^{\footnotemark}$\footnotetext{\url{https://github.com/numenta/nupic-legacy}} legacy framework, implemented in Python 2.7. This dependency constrained the entire software stack to outdated technologies, including ROS 1 Melodic and Ubuntu 18.04, limiting maintainability, scalability, and long-term support.

This new implementation of NeoSLAM $^{\footnotemark}$\footnotetext{\url{https://github.com/NeuroscienceRobotics/neoslam}} overcomes the aforementioned limitations. In summary, the contributions of this work are as follows:

\begin{itemize}
    \item \textbf{Architectural Refactoring:} The original codebase was fully refactored into a modular architecture, with a redesigned execution pipeline that significantly improves computational efficiency. By reducing the processing time per iteration, this refactoring substantially mitigates input data loss, enabling the system to operate at near-full capacity in real time with minimal discarding of incoming sensory data.

    \item \textbf{Software Stack Upgrade:} From an engineering standpoint, two foundational updates were carried out to support the objectives above and ensure practical applicability in real-world systems. First, the legacy \texttt{NuPIC} framework in Python 2.7 was replaced by the modern \texttt{htm.core} $^{\footnotemark}$\footnotetext{\url{https://github.com/htm-community/htm.core}} library, implemented in C++17 with Python 3 bindings, providing improved performance. This transition, in turn, enabled the migration to a ROS 2 Rolling-based architecture, which inherently provides modern middleware capabilities, improved communication mechanisms, and long-term maintainability.

    \item \textbf{Experimental Evaluation:} NeoSLAM is systematically evaluated against RatSLAM across three distinct datasets, providing a comprehensive comparative analysis.
\end{itemize}

\section{NeoSLAM Architecture: From Original Design to A New Implementation}\label{sec:meth01}

\begin{figure*}[]
\centerline{\includegraphics[width=.95\linewidth]{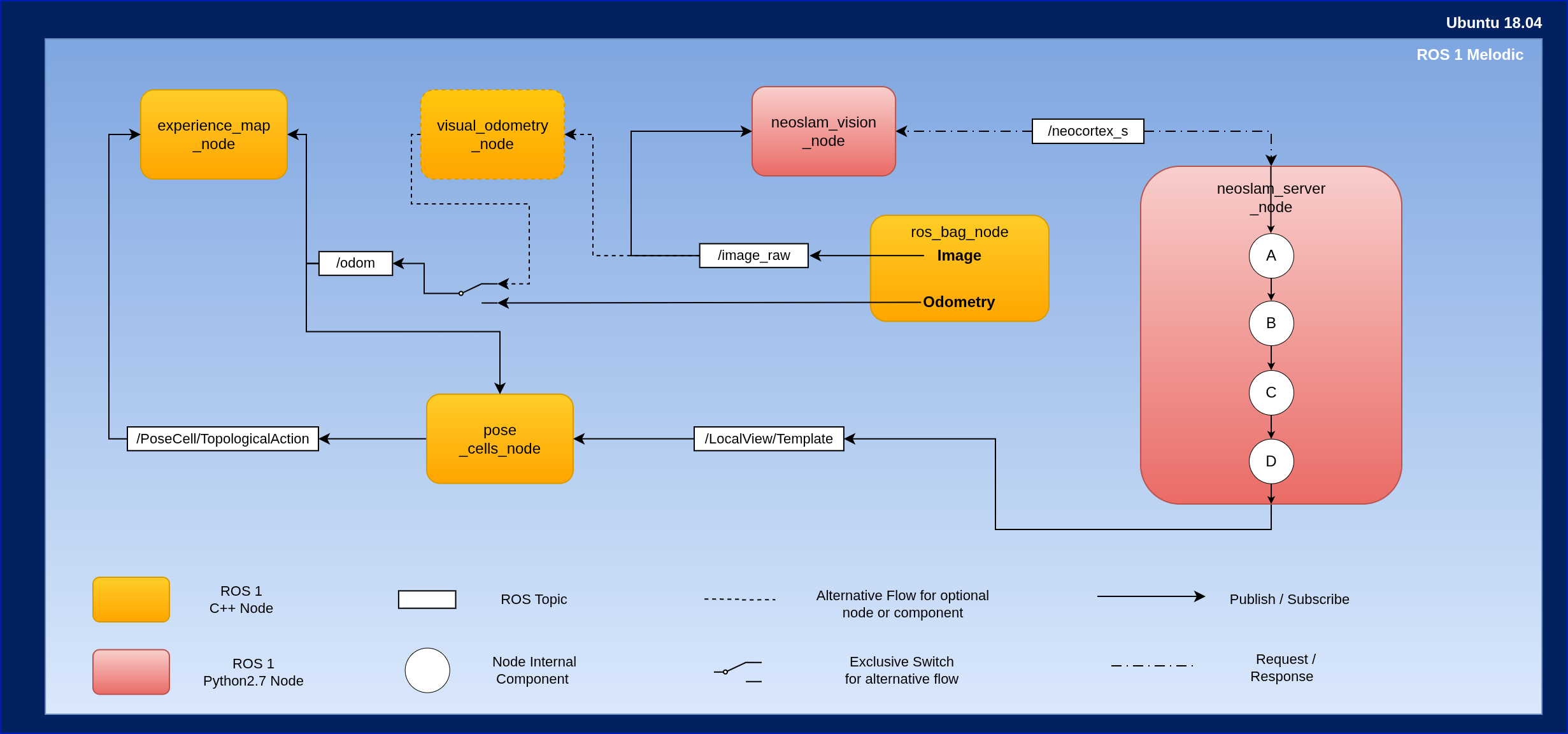}}
\caption{Original NeoSLAM architecture.}
\label{original_neoslam_arch}
\end{figure*}

The original NeoSLAM implementation by \cite{neoslam2024}, available at \url{https://github.com/cappizzino/neoslam_ws}, is illustrated in Figure~\ref{original_neoslam_arch}. In this architecture, each node --- whether implemented in C++ or Python --- runs as an independent operating system process, enabling inherent parallelism. Communication between nodes is carried out through ROS topics, which can follow one of two paradigms: publish/subscribe, a unidirectional mechanism in which a node broadcasts data to any number of subscribers, or request/response, a synchronous mechanism in which the requesting node blocks until a reply is received. Both paradigms are native to the ROS framework.

For the purpose of this work, regarding the proposal of a new architecture, the focus is strictly on the image data flow from the input sensor, received through the \verb|/image_raw| topic, to its transformation into a visual template, published on the \verb|/LocalView/Template| topic. In this flow, input images are read by the \texttt{neoslam\_vision\_node} at a frequency $f_{\text{in}}$. Upon receiving each frame, the \texttt{neoslam\_vision\_node} forwards it to the \texttt{neoslam\_server\_node} via a request on the \verb|/neocortex_s| topic. The \texttt{neoslam\_server\_node} is then responsible for processing the image, publishing the resulting visual template, and sending a response back to \texttt{neoslam\_vision\_node} to signal that processing is complete and that a new image may be submitted.

Upon receiving an image, the \texttt{neoslam\_server\_node} processes it sequentially through four internal operations, illustrated in figure~\ref{original_neoslam_arch} as components A through D: (A) image pre-processing and visual feature extraction via a convolutional neural network; (B) dimensionality reduction and binarisation of the extracted features; (C) generation of spatio-temporal visual templates through temporal memory processing; and (D) search and comparison against a history of visual templates for place recognition.

Let $\tau_A$, $\tau_B$, $\tau_C$, and $\tau_D$ denote the execution times of operations A, B, C, and D, respectively. The total processing time per frame within the \texttt{neoslam\_server\_node} is then given by $\tau = \tau_A + \tau_B + \tau_C + \tau_D$. For this architecture to process every incoming frame without data loss, $\tau$ must not exceed the input period $\tau_{\text{in}} = 1/f_{\text{in}}$, that is:
\begin{equation}
    \tau \leq \tau_{\text{in}}.
\end{equation}

In practice, however, the synchronous request/response mechanism further imposes that the \texttt{neoslam\_vision\_node} remains blocked during the entire duration of $\tau$, discarding any frames that arrive in the meantime. To illustrate the impact of this limitation, consider hypothetical execution times of $\tau_A = 50$\,ms, $\tau_B = 280$\,ms, $\tau_C = 20$\,ms, and $\tau_D = 350$\,ms, yielding a total processing time of $\tau = 700$\,ms, which corresponds to a maximum sustainable input frequency of approximately $1.43$\,Hz. For an input stream at $f_{\text{in}} = 1$\,Hz ($\tau_{\text{in}} = 1000$\,ms), the condition $\tau \leq \tau_{\text{in}}$ is satisfied and no data loss occurs. However, for a more demanding input rate of $f_{\text{in}} = 10$\,Hz ($\tau_{\text{in}} = 100$\,ms), the system is only capable of processing one out of every seven frames, resulting in a data loss of approximately $85.7$\%.
Consequently, for many real-time applications in which the input image frequency exceeds what this architecture can sustain, a significant fraction of incoming data is lost without being processed.

\begin{figure*}[]
\centerline{\includegraphics[width=.95\linewidth]{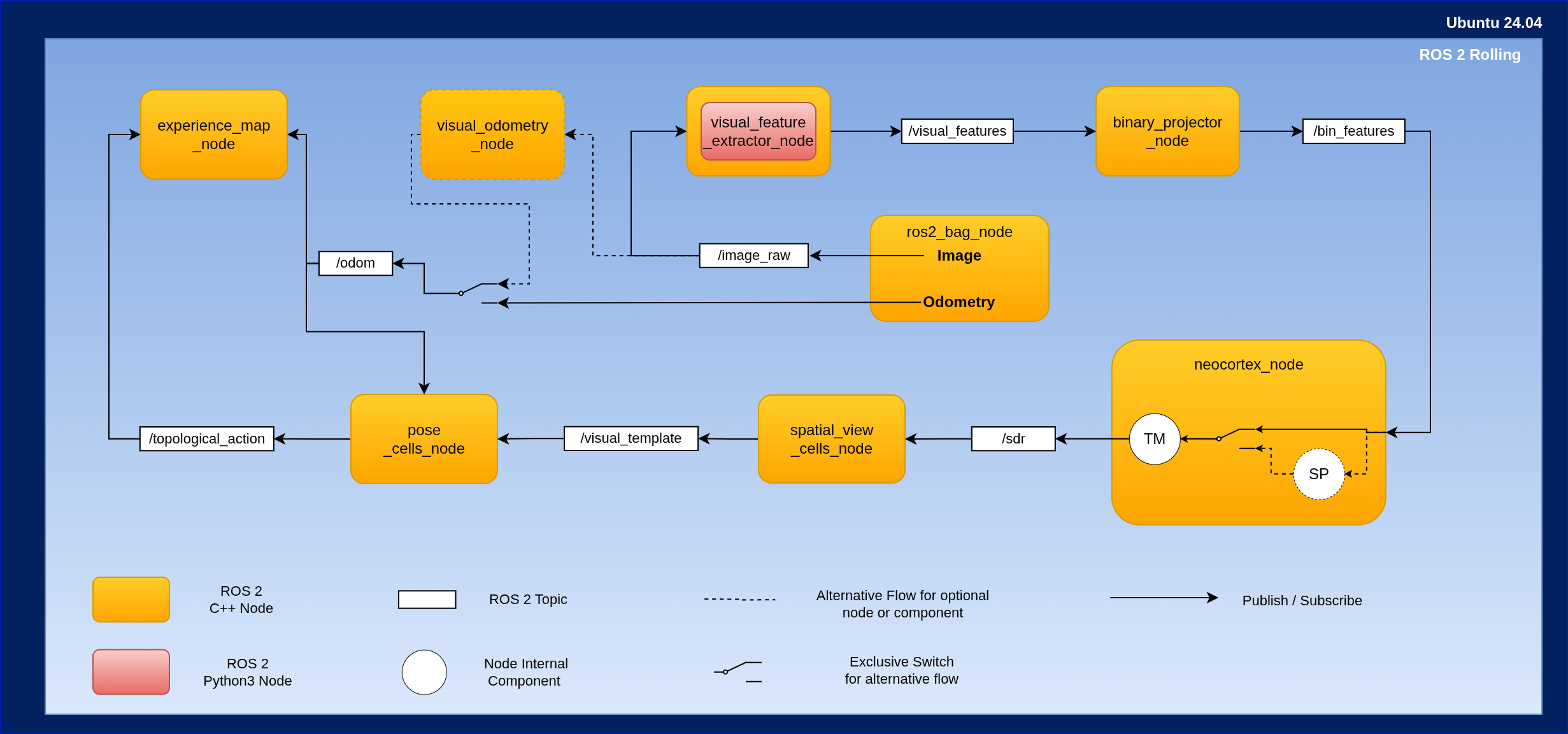}}
\caption{New NeoSLAM architecture.}
\label{neoslam_arch}
\end{figure*}

To address this limitation, the proposed architecture decomposes the monolithic \texttt{neoslam\_server\_node} into four independent nodes, \texttt{visual\_feature\_extractor\_node}, \texttt{binary\_projector\_node}, \texttt{neocortex\_node} and \texttt{spatial\_view\_cells\_node}, each corresponding to one of the sequential operations A through D. The new architecture is ilustrated in figure \ref{neoslam_arch}. Since each node operates as a separate process and communicates through unidirectional publish/subscribe topics, the pipeline becomes fully concurrent. Under this scheme, the maximum sustainable input frequency of the overall pipeline is no longer constrained by the total latency $\tau$, but rather by the slowest individual stage, known as the bottleneck. The maximum throughput of each node is given by $f_i = 1/\tau_i$, and the pipeline throughput is:
\begin{equation}
    f_{\text{pipeline}} = \min\left(\frac{1}{\tau_A},\, \frac{1}{\tau_B},\, \frac{1}{\tau_C},\, \frac{1}{\tau_D}\right).
\end{equation}
Using the same hypothetical values, the per-node maximum frequencies are $f_A = 20$\,Hz, $f_B \approx 3.57$\,Hz, $f_C = 50$\,Hz, and $f_D \approx 2.86$\,Hz, yielding a pipeline throughput of $f_{\text{pipeline}} \approx 2.86$\,Hz — a twofold improvement over the original architecture's $1.43$\,Hz, achieved without any modification to the individual processing operations themselves.

Beyond the architectural redesign, additional modifications of a more technological nature were also carried out, equally relevant given that the ultimate goal is to propose an algorithm suitable for real-time applications. The original nodes, implemented in Python 2.7, were ported to C++, a transition that required replacing the legacy \texttt{NuPIC} framework (\url{https://github.com/numenta/nupic-legacy}) with the modern \texttt{htm.core} library (\url{https://github.com/htm-community/htm.core}), implemented in C++17 with Python 3 bindings. The only exception is the AlexNet convolutional neural network, which was retained in Python 3 but invoked from C++ through a binding mechanism. Furthermore, the random matrix files were converted from \texttt{.NPY} to \texttt{.BIN} format, reducing memory usage and loading times, and the Eigen (\url{https://libeigen.gitlab.io/}) and CRoaring (\url{https://github.com/RoaringBitmap/CRoaring}) libraries were adopted for efficient matrix operations and sparse binary data handling, respectively. Collectively, these technological upgrades further reduced the per-node processing times from $\tau_A$, $\tau_B$, $\tau_C$, and $\tau_D$ to $\tau_A'$, $\tau_B'$, $\tau_C'$, and $\tau_D'$, respectively, where each updated value is considerably smaller than its original counterpart, further increasing the pipeline throughput beyond what the architectural restructuring alone could achieve. The extent of this data loss for both architectures are quantified and discussed in the section \ref{sec:results}.

\section{New NeoSLAM: System Overview}\label{sec:meth02}

The new implementation of NeoSLAM comprises seven main modules: Visual Feature Extractor, Binary Projector, Neocortex, Spatial-View Cells, Pose Cells, Experience Map and Visual Odometry. Figure \ref{neoslam_arch} illustrates the main modules described as ROS 2 nodes, along with the corresponding ROS topics describing the data flow between them.


\subsection{Visual Feature Extractor Node}

The Visual Feature Extractor Node is responsible for receiving input images, selecting frames at a defined temporal stride, extracting a region of interest, and computing high-level visual features through a Convolutional Neural Network.

This version introduces the \verb|frame_stride| parameter $s \in \mathbb{N}$, which controls the temporal resolution of the input stream. Given a sequence of input images, only one
out of every $s$ consecutive frames is forwarded for processing, while the remaining $s-1$ frames are discarded. This uniform temporal downsampling reduces computational load prior to feature extraction and provides a mechanism to avoid processing temporally redundant frames according to a particular application. Following frame selection, cropping parameters are applied in the pre-processing stage to restrict the analysis to relevant visual regions. Let the downsampled input image be represented as $I \in \mathbb{R}^{W \times H}$, where $W$ and $H$ denote width and height, respectively. When \texttt{crop\_image} is enabled, a subregion is defined by horizontal limits $[x_{\min}, x_{\max})$ and vertical limits $[y_{\min}, y_{\max})$, satisfying $0 \le x_{\min} < x_{\max} \le W$ and $0 \le y_{\min} < y_{\max} \le H$.
The cropped image is then given by
\begin{equation}
I_{\text{crop}}(x',y') = I(x' + x_{\min},\, y' + y_{\min}),
\end{equation}
for $0 \le x' < x_{\max}-x_{\min}$ and $0 \le y' < y_{\max}-y_{\min}$.
This operation ensures that feature extraction is restricted to the selected region of interest.

Subsequently, a pre-trained \textit{AlexNet conv3} feature extractor \cite{neubert2019, neoslam2024} produces the descriptor
$\mathbf{f} = \text{CNN}(I_{crop})$
where $\mathbf{f} \in \mathbb{R}^{n}$ with $n = 64896$. The descriptor is published to the \verb|/visual_features| topic.

\subsection{Binary Projector Node}

The Binary Projector Node is composed of two stages: dimensionality reduction and binarization.

First, the dimensionality reduction stage is performed using a random gaussian projection defined as $\mathbf{d} = P\mathbf{f}$, where $P \in \mathbb{R}^{m \times n}$ is the gaussian random projection matrix, and $\mathbf{d} \in \mathbb{R}^{m}$ is the reduced descriptor with $m=1024$ dimensions, with $m < n$. This reduction decreases computational cost and memory usage while maintaining structural properties necessary for subsequent encoding \cite{bingham2001}.

In the second stage, the Sparse Locality Sensitive Binary Hashing (sLSBH) method \cite{neubert2019, neoslam2024} performs binarization and sparsification to obtain a compact and robust binary representation. Given the projected descriptor $\mathbf{d} \in \mathbb{R}^{m}$, with $m = 1024$ in our configuration, the algorithm selects the $s = 25\%$ largest and $s = 25\%$ smallest components of $\mathbf{d}$ to construct a sparse binary code.

The final binary vector is defined as the concatenation
\begin{equation}
\mathbf{z} = \left[ \mathbf{z}^{+} \; \mathbf{z}^{-} \right]^T ,
\end{equation}
where $\mathbf{z}^{+} \in \{0,1\}^{m}$ encodes the largest coefficients and $\mathbf{z}^{-} \in \{0,1\}^{m}$ encodes the smallest coefficients of $\mathbf{d}$. Since $m = 1024$ and $s = 25\%$, each vector contains $\frac{s}{100}\cdot m = 256$ active bits.

Therefore, each subvector $\mathbf{z}^{+}$ and $\mathbf{z}^{-}$ has length $1024$ with $256$ active bits, and their concatenation produces the final descriptor $\mathbf{z} \in \{0,1\}^{2m}$ of dimension $2048$ with a total of $512$ active bits. The $\mathbf{z}$ binary descriptor is written in the \verb|/bin_features| topic.

\subsection{Neocortex Node}

The Neocortex Node transforms the reduced binary visual information into a format that captures spatial and temporal features grounded in the principles of \textit{Hierarchical Temporal Memory} (HTM) \cite{htm2016}. To facilitate understanding of this module, the fundamental structures required for operating an HTM network are described: Sparse Distributed Representation (SDR), Spatial Pooler (SP), and Temporal Memory (TM).


\subsubsection{Sparse Distributed Representation}

An SDR is defined as an $n$-dimensional binary vector $\mathbf{x} = [b_1, \ldots, b_{n}]$,
where $\mathbf{x} \in \{0,1\}^{p}$ and only a small fraction of components are active ($b_i = 1$).

The overlap $\Phi$ between two SDRs $\mathbf{x_1}$ and $\mathbf{x_2}$ measures similarity by counting shared active bits:
\begin{equation}
\Phi(\mathbf{x_1}, \mathbf{x_2}) = \mathbf{x_1} \cdot \mathbf{x_2} = \|\mathbf{x}_1 \cap \mathbf{x}_2\|_0.
\label{eq:sdr_overlap}
\end{equation}

A match $\Psi$ occurs when the overlap exceeds a predefined threshold $\beta$:
\begin{equation}
\Psi = \Phi(\mathbf{x_1}, \mathbf{x_2}) \geq \beta.
\label{eq:sdr_match}
\end{equation}

\subsubsection{Spatial Pooler}

The Spatial Pooler (SP) converts inputs into SDRs, providing properties that are typically absent in raw inputs. By activating only a small fraction of mini-columns through a competitive mechanism, the SP enforces fixed sparsity and improves representational efficiency. 
At the same time, it increases robustness to noise by mapping similar inputs to overlapping sets of active columns, producing stable representations under small input perturbations \cite{cui2017spatialpooler}. 
In this work, the binary descriptor is directly converted to an SDR and the SP component is not used.

\subsubsection{Temporal Memory}

The Temporal Memory (TM) is a recurrent binary-state network that accumulates information over consecutive input SDRs in order to encode the temporal context of the current input. It operates on a layer composed of $n$ minicolumns ($C_j$), each containing $m$ cells ($c_{i,j}$), where contextual information is represented by distinct cells within active minicolumns. The TM receives as input a SDR and, at each time step $t \in \mathbb{N}_0$, its state is characterized by three binary variables per cell: active $a_{i,j}^t$, predictive (depolarized) $\pi_{i,j}^t$, and winner $w_{i,j}^t$, grouped into the matrices $A^t$, $\Pi^t$, and $W^t$, respectively.

The TM dynamics begin with the calculation of the active state of the cells according to the equation (Eq. \ref{eq:active_state}):
\begin{equation}
a_{i,j}^t = 1 \iff A_j^t \land \left( \pi_{i,j}^{t-1} \lor \left( \forall m : \neg \pi_{m,j}^{t-1} \right) \right)
\label{eq:active_state}
\end{equation}
A cell is activated within a winning column ($A_j^t=1$) if it was previously predicted ($\pi_{i,j}^{t-1}=1$) or if no cells in that minicolumn were in a predictive state. In the latter case, a ``bursting'' phenomenon occurs, where all cells in the column are activated to represent a new context.

Let $D_{ij}^d=\{s_{ij}^d\}$ be the synaptic permanence matrix from a particular distal segment $d$ where each synapse is bounded by $0 \le s_{i,j}^d \le 1$, and $\tilde{D}_{ij}^d=1$ if ${D_{ij}^d>\epsilon}$ be the binary matrix of connected synapses for some permanence threshold. Then, the predictive state for the current time step $t$ is determined by the equation \ref{eq:predictive_state}:
\begin{equation}
\pi_{i,j}^t = 1 \iff \exists_d \left\| \tilde{D}_{i,j}^d \circ A^{t} \right\|_1 > \theta
\label{eq:predictive_state}
\end{equation}
where $\theta \in \mathbb{N}$ represents the dendritic segment activation threshold and $\circ$ denotes the element-wise multiplication (Schur product) . A cell becomes depolarized if at least one of its distal dendritic segments $d$ has a sufficient number of connected synapses with currently active presynaptic cells.

Learning occurs through the selection of winner cells that reinforce sequence transitions. If a prediction is successful, the winner cell is identified by equation \ref{eq:winner_pred}:
\begin{equation}
\begin{aligned}
w_{i,j}^t = 1 \iff A_j^t \wedge (\pi_{i,j}^{t-1} > 0) \\
\wedge \|\tilde{D}_{i,j}^d \circ A^{t-1}\|_1 > \theta
\end{aligned}
\label{eq:winner_pred}
\end{equation}

However, if bursting occurs, Equation \ref{eq:winner_burst} selects the cell that was closest to activation (the one with the highest $L_1$ norm), even if it was below the threshold $\theta$, to represent the future context:
\begin{equation}
\begin{aligned}
w_{i,j}^t = 1 \Leftrightarrow & A_j^t \wedge \left( \forall m : \neg \pi_{m,j}^{t-1} \right) \\
& \wedge \|\dot{D}_{i,j}^d \circ A^{t-1}\|_1 = \\
& \max_i \left( \|\dot{D}_{i,j}^d \circ A^{t-1}\|_1 \right)
\end{aligned}
\label{eq:winner_burst}
\end{equation}
where $\dot{\mathbf{D}}_{ij}^{d} = 1$ if $\mathbf{D}_{ij}^{d} > 0$, and $0$ otherwise.

Finally, synaptic permanence values are adjusted using a Hebbian-like rule (Eq. \ref{eq:permanence}):
\begin{equation}
\Delta D_{i,j}^d = \epsilon_i \left( \dot{D}_{i,j}^d \circ A^{t-1} \right) - \epsilon_d \dot{D}_{i,j}^d
\label{eq:permanence}
\end{equation}
where the factors $\epsilon_i$ and $\epsilon_d$ increase the permanence of synapses with active presynaptic cells and decrease it for those with inactive cells, respectively.

Now that the fundamental structures of HTM network were defined, the specific parameters implemented in this work are explained. 
The TM module inside the Neocortex node receives an input SDR $\mathbf{z} \in \{0,1\}^{2048}$ containing 512 active bits (activating 512 columns $A_j^t$). Through the HTM dynamics, this representation is expanded into a higher-dimensional SDR $\mathbf{y} \in \{0,1\}^{32 \cdot 2048} = \{0,1\}^{65536}$, corresponding to 32 cells per minicolumn. Each column represents distinct visual features and while all cells inside a column can be activated by the same SDR input, the temporal context distingshes them. Therefore, the resulting output SDR increases representational capacity, enabling robust sequence encoding and improved discrimination of temporally correlated visual patterns.

\subsection{Spatial-View Cells Node}

In rodents, navigation is largely described as occurring from place to place and relies primarily on local tactile and olfactory cues. 
In contrast, the highly developed visual system of primates enables the identification of locations directly from viewed scenes, a distinction that has important implications for how the primate, including the human hippocampus, supports episodic memory processes \cite{spatialview2025}. Inspired by Spatial-View cells found in primates, a computational model aimed at visual place recognition and loop closure detection was proposed \cite{neoslam2024}, associating neocortical visual representations with previously stored visual experiences to recognize revisited locations. The Spatial-View Cells Node introduced in this work implements the same underlying model with improved computational efficiency and presents it using a complete mathematical formulation.

At each time step $t \in \mathbb{N}_0$, the system observes a SDR $\mathbf{y}_t \in \{0,1\}^{65536}$ generated by the neocortical model, with 512 active bits.
Those observations are grouped into intervals $I_k = [t_k^{start},t_k^{end}]$, where $k$ denotes the interval index and $\rho_k = t_k^{end}-t_k^{start} \geq 1$ the interval size. Each interval maintains an accumulated SDR representation defined as the union of all SDRs within the interval 
\begin{equation}
\mathbf{Y}_k = \bigcup_{t=t_k^{start}}^{t_k^{end}} \mathbf{y}_t.
\end{equation}

Given a new observation $\mathbf{y}_t$, two parameters define if the observation will join the current interval or create a new one: the similarity threshold $\theta_\alpha \in \mathbb{N}^+$ for interval membership (\verb|theta_alpha|) and the maximum interval size threshold $\theta_\rho \in \mathbb{N}^+$ (\verb|theta_rho|). If the similarity (which corresponds to the overlap equation \ref{eq:sdr_overlap}) between the current interval $k_{curr}$, computed as
\begin{equation}
\alpha(t) = \mathbf{Y}_{k_{curr}} \cdot \mathbf{y}_t = \|\mathbf{Y}_{k_{curr}} \cap \mathbf{y}_t\|_0
\end{equation}
is high enough $\alpha(t) \ge \theta_\alpha$ and does not extrapolate the maximum interval size $\rho_{k_{curr}} < \theta_\rho$, then the interval is extended $\mathbf{Y}_{k_{curr}} \leftarrow \mathbf{Y}_{k_{curr}} \cup \mathbf{y}_t$; otherwise, a new interval is created and its representation is initialized as $\mathbf{Y}_{k+1}=\mathbf{y}_t$.
All interval representations are stored in the matrix
$
\mathbf{M}_{intervals}^{T} =
\begin{bmatrix}
\mathbf{Y}_1^{T} &
\mathbf{Y}_2^{T} &
\cdots &
\mathbf{Y}_{K}^{T}
\end{bmatrix}
$
where $K$ is the total number of intervals.

Loop closures are detected by computing the similarity between the current observation and all intervals, producing the vector $\mathbf{s}(t) = [s_1(t),\dots,s_{K}(t)]^T$, where $s_k(t) = \mathbf{Y}_k \cdot \mathbf{y}_t = \|\mathbf{Y}_k \cap \mathbf{y}_t\|_0.$
Here, two other parameters are important for considering loop closures: the loop closure threshold $\theta_\sigma \in \mathbb{N}^+$ used to define the loop closure occurrence and the new proposed parameter \verb|exclude_recent_intervals| defined as $\omega \in \mathbb{N}_0$ used to avoid trivial matches with recent observations. Recent intervals are excluded by considering only the first $K-\omega$ elements of $\mathbf{s}(t)$. Candidate loop closures are then defined as $\mathcal{M}(t)=\{k \in \{1,\dots,K-\omega\}: s_k(t)>\theta_\sigma\}.$
If $\mathcal{M}(t)\neq\emptyset$, the best matching interval is selected as $k^*=\arg\max_k s_k(t)$.
Each interval is associated with a visual template identifier through the mapping $\phi:\{1,\dots,K\}\rightarrow\mathbb{N}$. The template assigned to the current observation is
\begin{equation}
\tau(t)=
\begin{cases}
\phi(k^*) & \text{if } \mathcal{M}(t)\neq\emptyset \\
\tau_{curr} & \text{if } \mathcal{M}(t)=\emptyset \land k_{prev}=k_{curr} \\
\tau_{next} & \text{if } \mathcal{M}(t)=\emptyset \land k_{prev} \neq k_{curr}
\end{cases}
\end{equation}

Thus, loop closures reuse the template of the most similar interval, while new intervals without matches generate new template identifiers. The template message is published through the \verb|/visual_template| topic.

The last three nodes (Pose Cells, Experience Map, and Odometry Node) remain unchanged from RatSLAM. A full description of these models can be found in \cite{ratslam2004} and \cite{icar2025}.

\section{Experimental Results}\label{sec:results}


This section presents results from experiments on three datasets: Robotarium and iRat Australia (terrestrial), and FIU MMC Lake (aquatic, collected by an Unmanned Surface Vehicle). The results are divided into two parts: Part~I follows the methodology of Section~\ref{sec:meth01}, comparing the original and proposed NeoSLAM architectures in terms of real-time performance and robustness; Part~II follows the methodology of Section~\ref{sec:meth02}, comparing the proposed NeoSLAM against RatSLAM in terms of topological map accuracy, measured by the Euclidean distance between the ground truth and the generated trajectories at each time sample.
Both NeoSLAM and RatSLAM rely on a large number of parameters. To support reproducibility and assist in system tuning, a complete list of parameters along with their default values is available on the project page $^{\footnotemark}$\footnotetext{\url{https://neurosciencerobotics.github.io/neoslam.github.io/}}.

\subsection{Robotarium Dataset}

\begin{figure}[t]
    \centering
    \includegraphics[width=.90\linewidth]{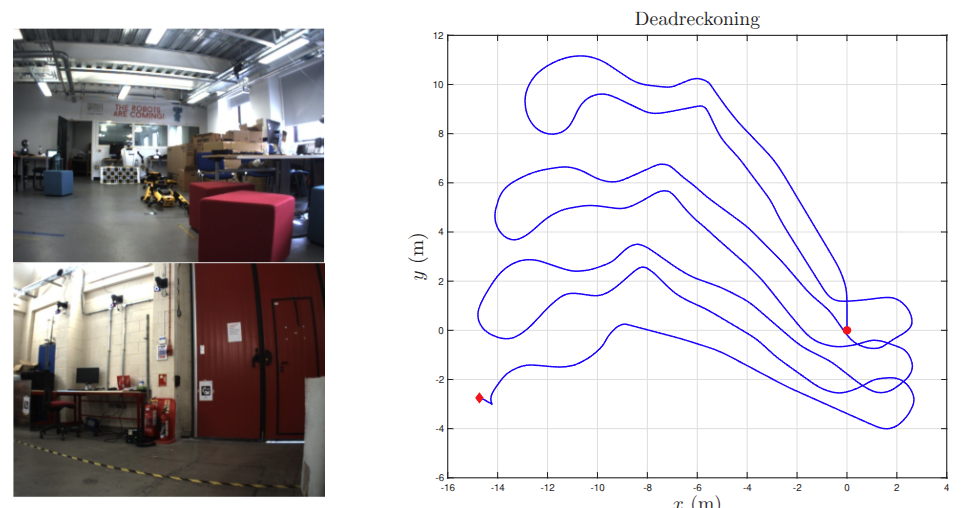}
    \caption{(a) Frames examples from the frontal camera (b) Odometry from Husky at Robotarium experiment.}
    \label{fig_Robotarium}
\end{figure}

This dataset was created at the Robotarium laboratory at Heriot-Watt University and was introduced in \cite{neoslam2024}. Although NeoSLAM has already been tested on this dataset, the parameters were not reported. A Husky A200 UGV (Unmanned Ground Vehicle) was used to generate and collect the data. Images were captured using only the left camera (\autoref{fig_Robotarium}-(a)), while odometry readings were obtained from the wheel encoders. A noticeable orientation error in the odometry between laps can be observed in \autoref{fig_Robotarium}-(b), as the trajectories would otherwise overlap. 

\subsubsection{Part I: A comparison between the Original NeoSLAM and the new NeoSLAM implementation}

\begin{figure}[t]
    \begin{minipage}[t]{0.48\linewidth}
        \centering
        \includegraphics[width=.9\linewidth]{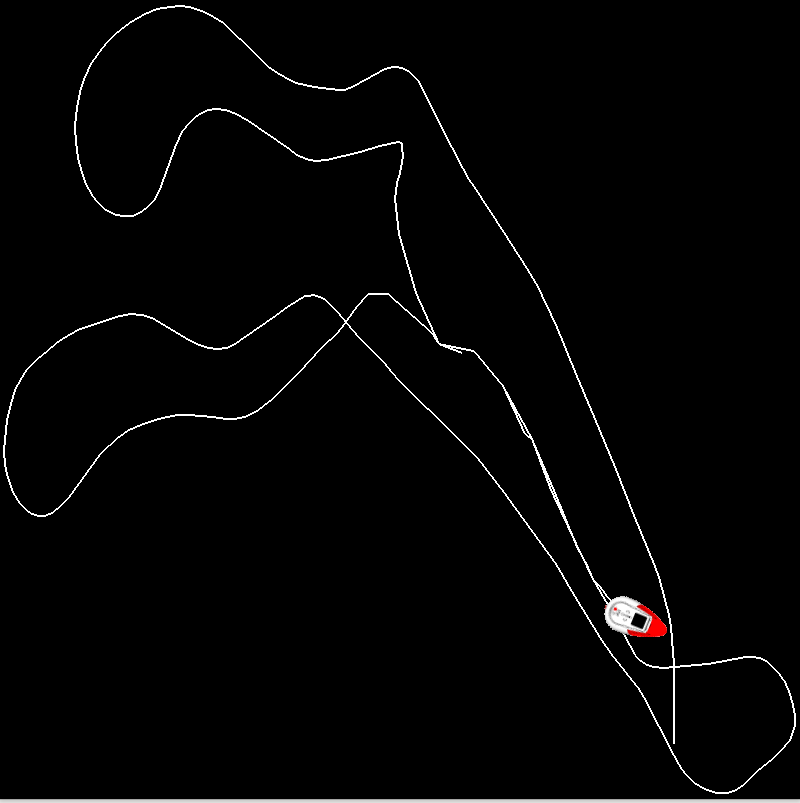}
    \end{minipage}
    \begin{minipage}[t]{0.48\linewidth}
        \centering
        \includegraphics[width=.9\linewidth]{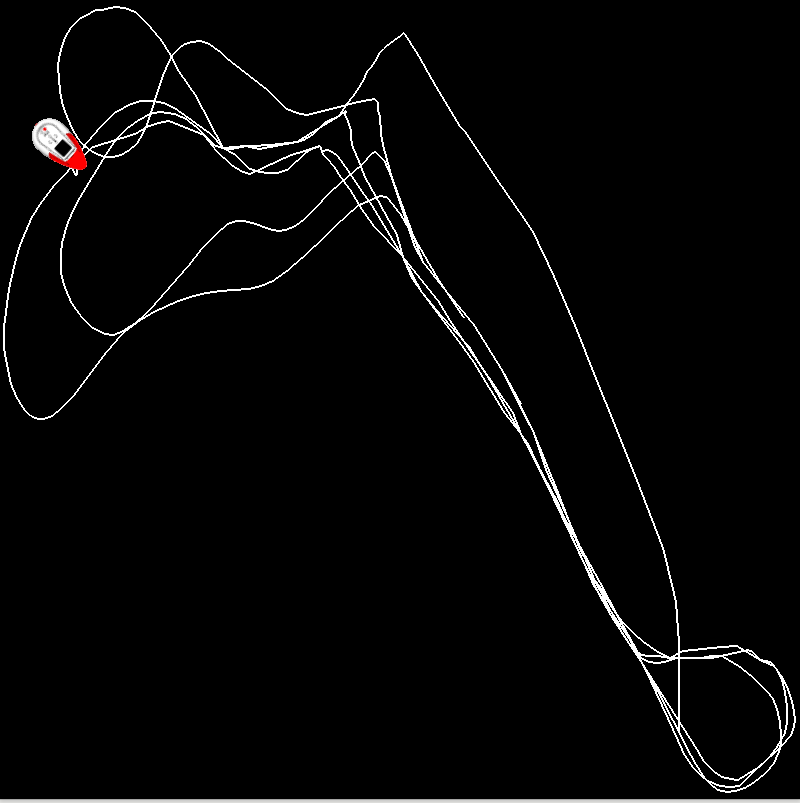}
    \end{minipage}
    \caption{Experience map evolution of Robotarium with the original NeoSLAM in real-time execution.}
    \label{fig_exp_map_evolution_robotarium}
\end{figure}

It is clear from the evolution of the experience map in Figure~\ref{fig_exp_map_evolution_robotarium} for the Robotarium dataset that the system fails on some loop closure opportunities, resulting in a degraded topological map compared to the ground truth. It is worth noting that the parameters used in this experiment with the original NeoSLAM are the same as those presented later for the proposed new NeoSLAM, so that the lack of robustness is attributed to the aggressive data loss observed in Table~\ref{tab:data_loss_robotarium}. Furthermore, it is important to emphasise that the results obtained here for the Robotarium dataset may differ from those reported by \cite{neoslam2024} due to the non-deterministic nature of the machine execution, which can introduce variability in timing and processing order across runs.

\begin{table}[ht]
\centering
\caption{Data loss comparison between Original NeoSLAM and New NeoSLAM during real-time execution for Robotarium Dataset.}
\label{tab:data_loss_robotarium}
\begin{tabular}{lcc}
\toprule
\textbf{Algorithm} & \textbf{Processed Frames} & \textbf{Data Loss (\%)} \\
\midrule
Original NeoSLAM & 537 & 91\% \\
New NeoSLAM      & 5423 & 1\% \\
\bottomrule
\end{tabular}
\vspace{0.5em}

\footnotesize\textit{Experimental conditions: input rate = 13\,FPS, total input frames = 5453, repetitions = 3, experiment duration = 424,8 (s).}
\end{table}

\subsubsection{Part II: A comparison between NeoSLAM and RatSLAM}

\begin{figure}[t]
    \begin{minipage}[t]{0.48\linewidth}
        \centering
        \includegraphics[width=1\linewidth]{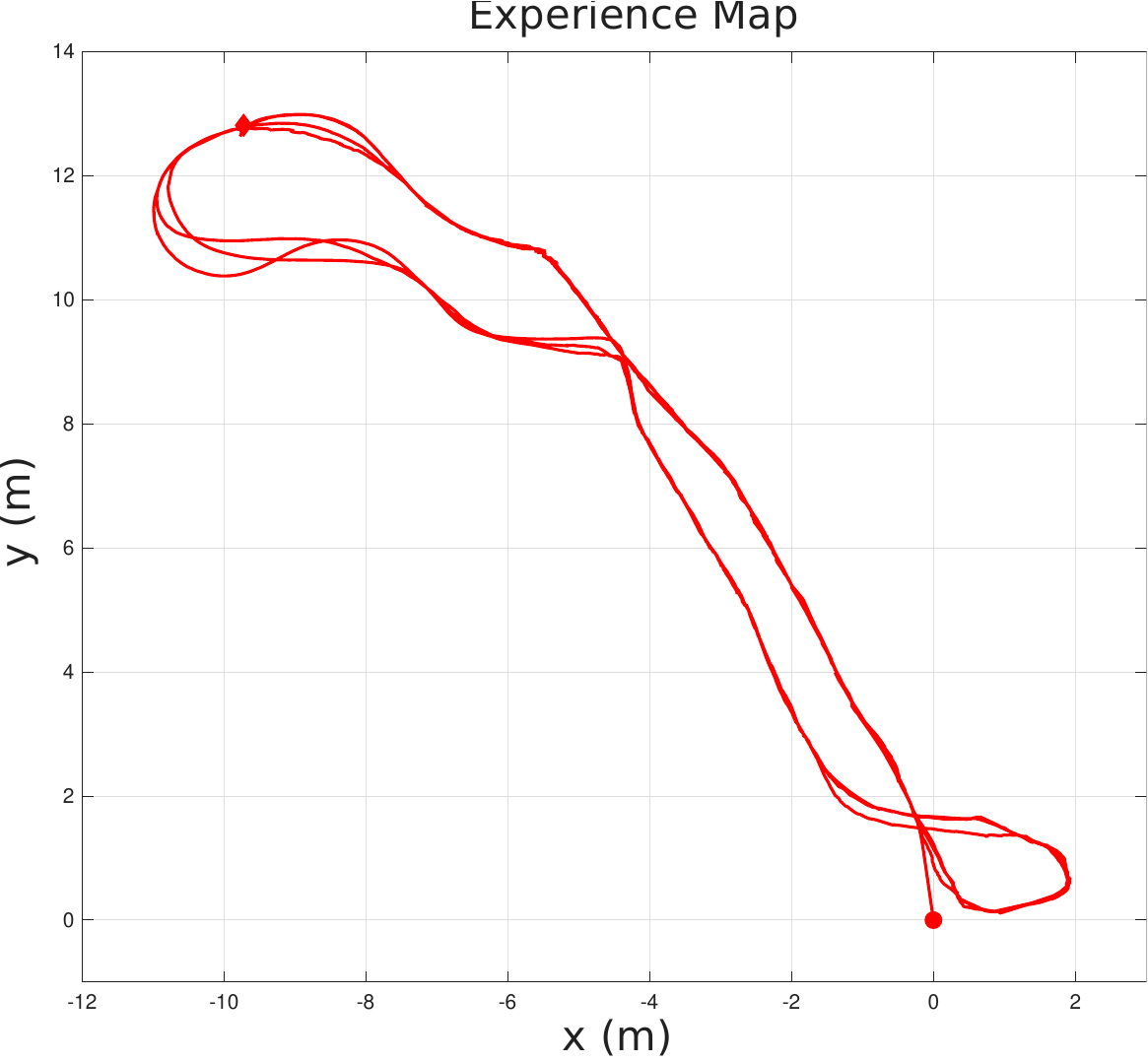}
    \end{minipage}
    \begin{minipage}[t]{0.48\linewidth}
        \centering
        \includegraphics[width=1\linewidth]{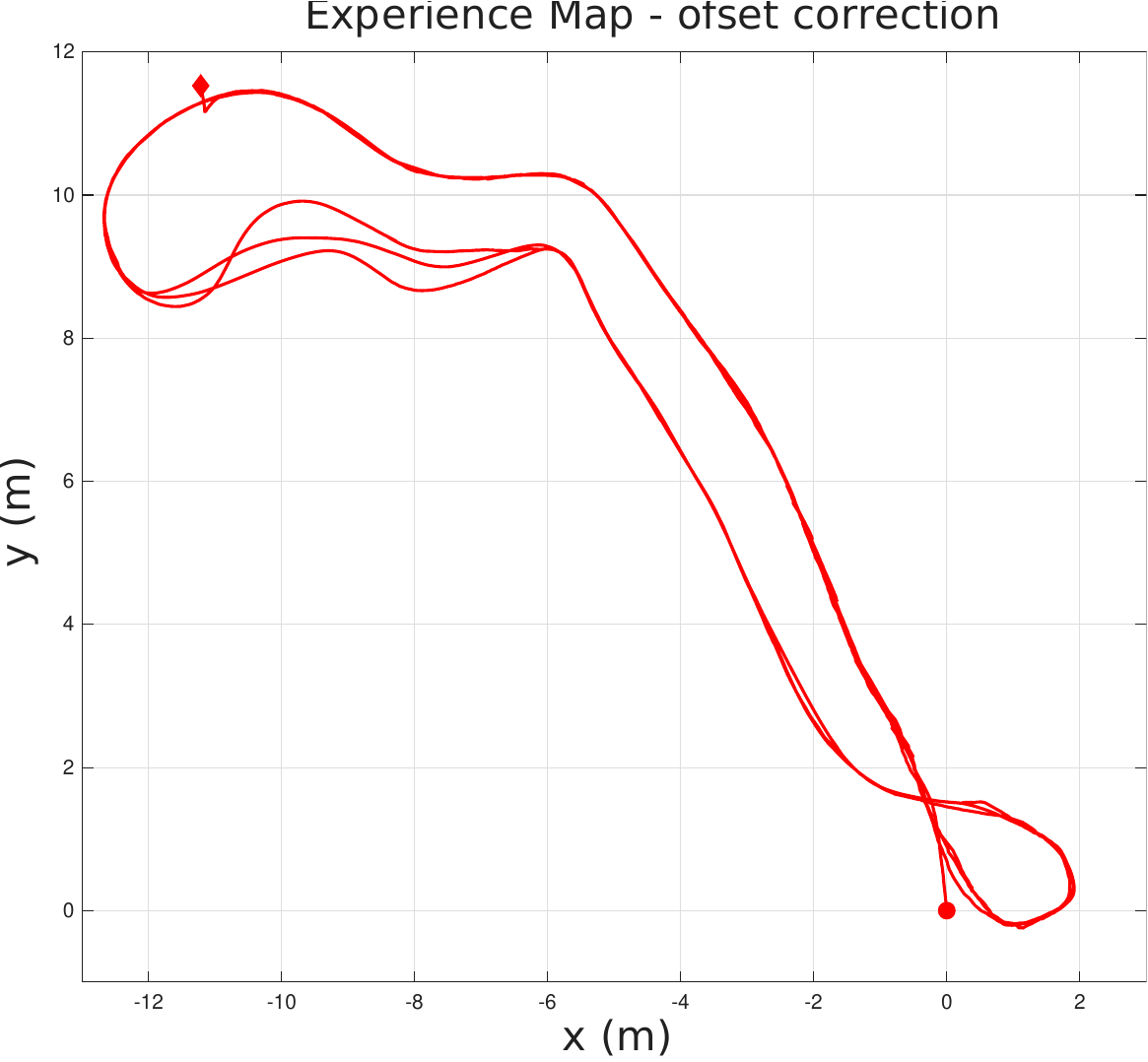}
    \end{minipage}
    \caption{Experience map for Robotarium dataset. (a) Result of RatSLAM algorithm and (b) NeoSLAM results.} 
    \label{fig_exp_map_robotarium}
\end{figure}

\autoref{fig_exp_map_robotarium} shows that both algorithms were able to generate consistent topological maps of the indoor environment. Despite visible differences between the trajectories, the lack of ground truth makes it impossible to judge which experience map is more accurate.

\subsection{iRat Australia Dataset}

\begin{figure}[b]
    \centering
    \includegraphics[width=1\linewidth]{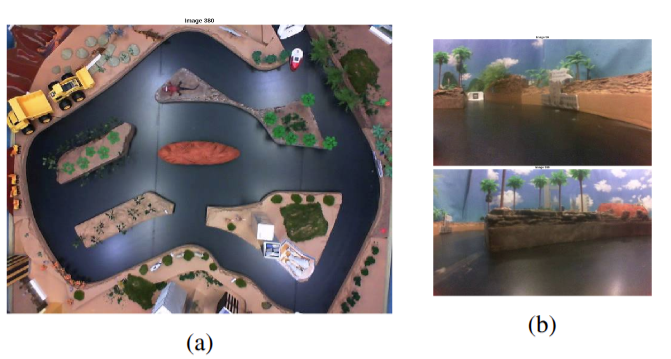}  
    \caption{iRat Australia: (a) Overhead camera and (b) frontal camera frames.} 
    \label{fig_irat2011}
\end{figure}

The iRat Australia dataset is presented in \cite{OpenRatSLAM2013}. The iRat robot has a differential wheel drive, a forward-facing camera, and wheel encoders, which provide odometric readings. A camera mounted overhead captured images that enabled the extraction of ground-truth information. \autoref{fig_irat2011} shows an overhead image and frontal camera samples.

\subsubsection{Part I: A comparison between the Original NeoSLAM and the new NeoSLAM implementation}

\begin{figure}[t]
    \begin{minipage}[t]{0.48\linewidth}
        \centering
        \includegraphics[width=.9\linewidth]{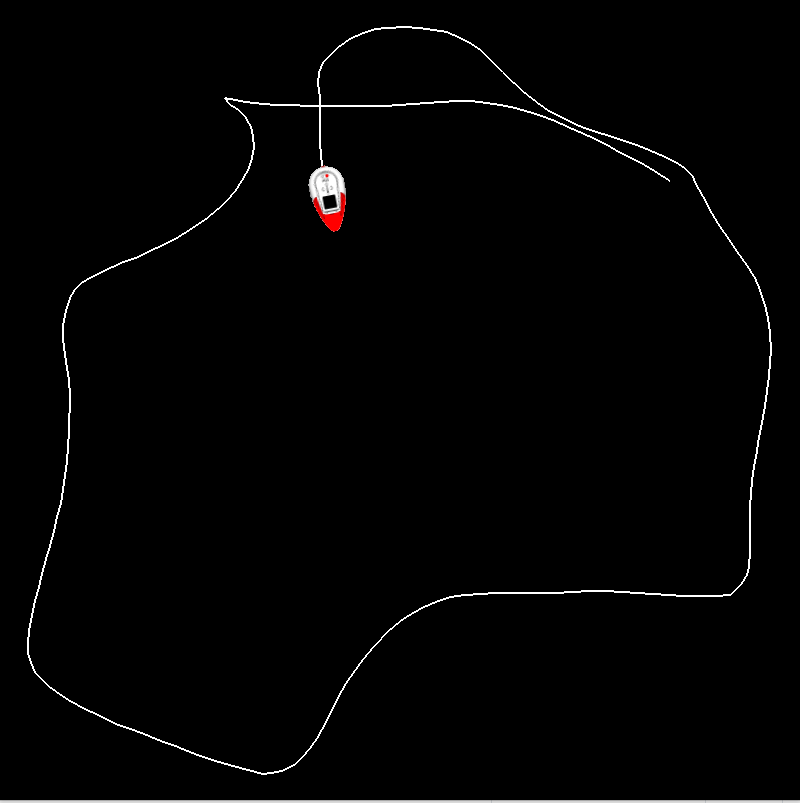}
    \end{minipage}
    \begin{minipage}[t]{0.48\linewidth}
        \centering
        \includegraphics[width=.9\linewidth]{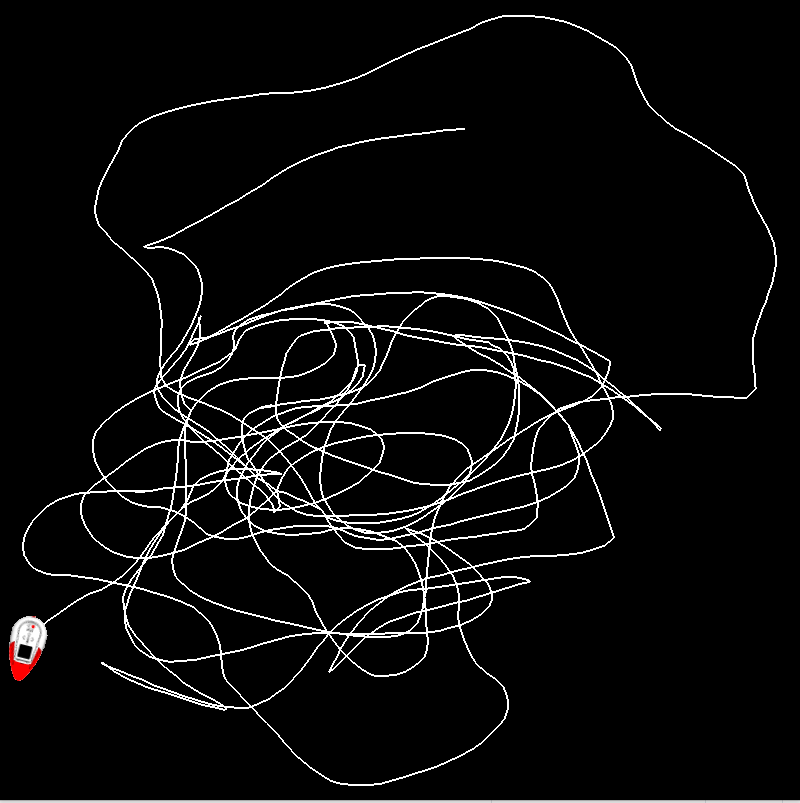}
    \end{minipage}
    \caption{Experience map evolution of iRat Australia with the original NeoSLAM in real-time execution.}
    \label{fig_exp_map_evolution_Irat}
\end{figure}

It is clear from the evolution of the experience map in Figure~\ref{fig_exp_map_evolution_Irat} for the iRat Australia dataset that the system fails on several loop closure opportunities, resulting in a topological map that is entirely incoherent with the ground truth. Again, the parameters used in this experiment with the original NeoSLAM are the same as those presented for the new NeoSLAM, so that the lack of robustness is attributed to the aggressive data loss observed in Table~\ref{tab:data_loss_irat}.

\begin{table}[ht]
\centering
\caption{Data loss comparison between Original NeoSLAM and New NeoSLAM during real-time execution for iRat Australia Dataset.}
\label{tab:data_loss_irat}
\begin{tabular}{lcc}
\toprule
\textbf{Algorithm} & \textbf{Processed Frames} & \textbf{Data Loss (\%)} \\
\midrule
Original NeoSLAM & 1532 & 90\% \\
New NeoSLAM      & 16657 & 0\% \\
\bottomrule
\end{tabular}
\vspace{0.5em}

\footnotesize\textit{Experimental conditions: input rate = 10\,FPS, total input frames = 16657, repetitions = 3, experiment duration = 15,9 (min).}
\end{table}

\subsubsection{Part II: A comparison between NeoSLAM and RatSLAM}

\begin{figure}[t]
    \begin{minipage}[t]{0.48\linewidth}
        \centering
        \includegraphics[width=1\linewidth]{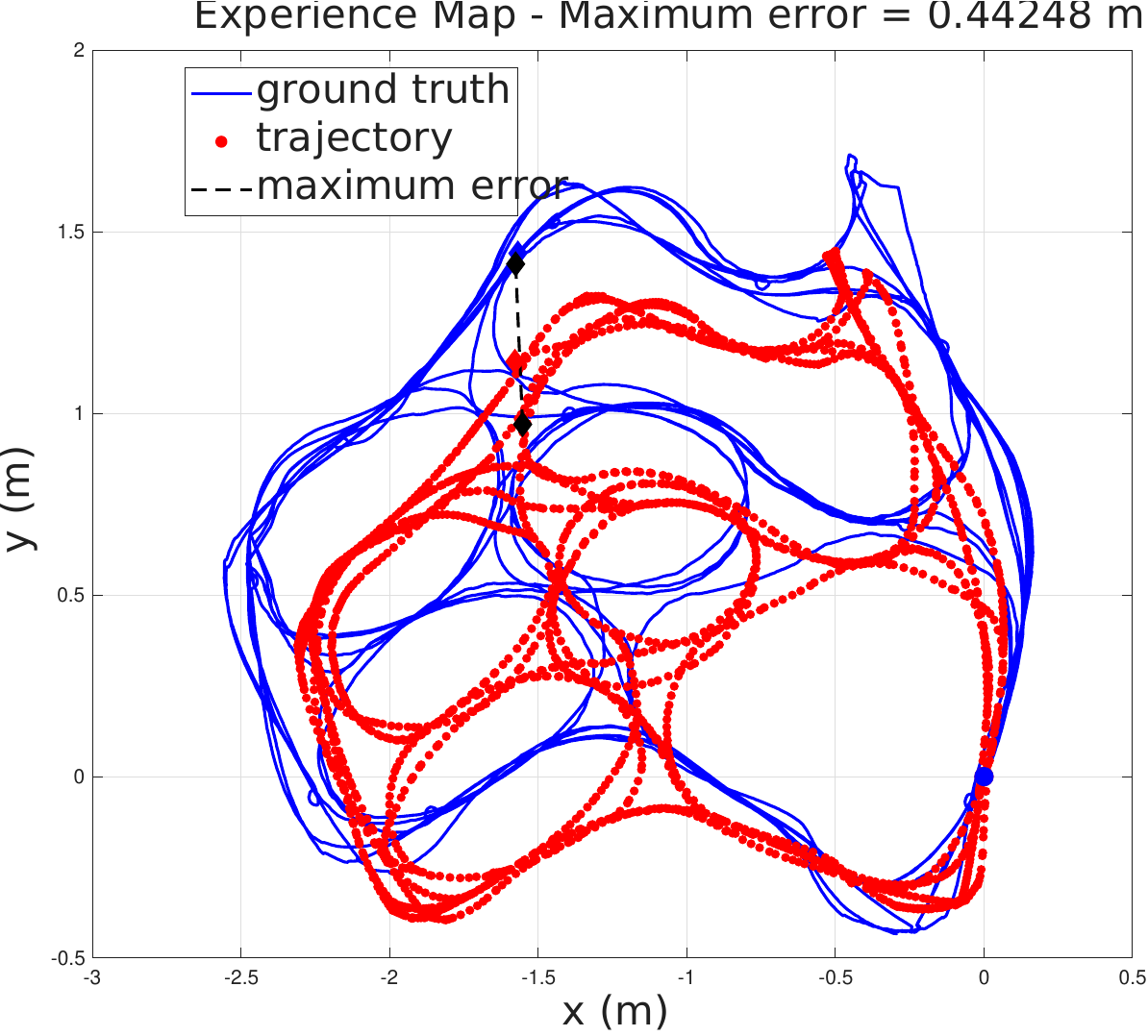}
    \end{minipage}
    \begin{minipage}[t]{0.48\linewidth}
        \centering
        \includegraphics[width=1\linewidth]{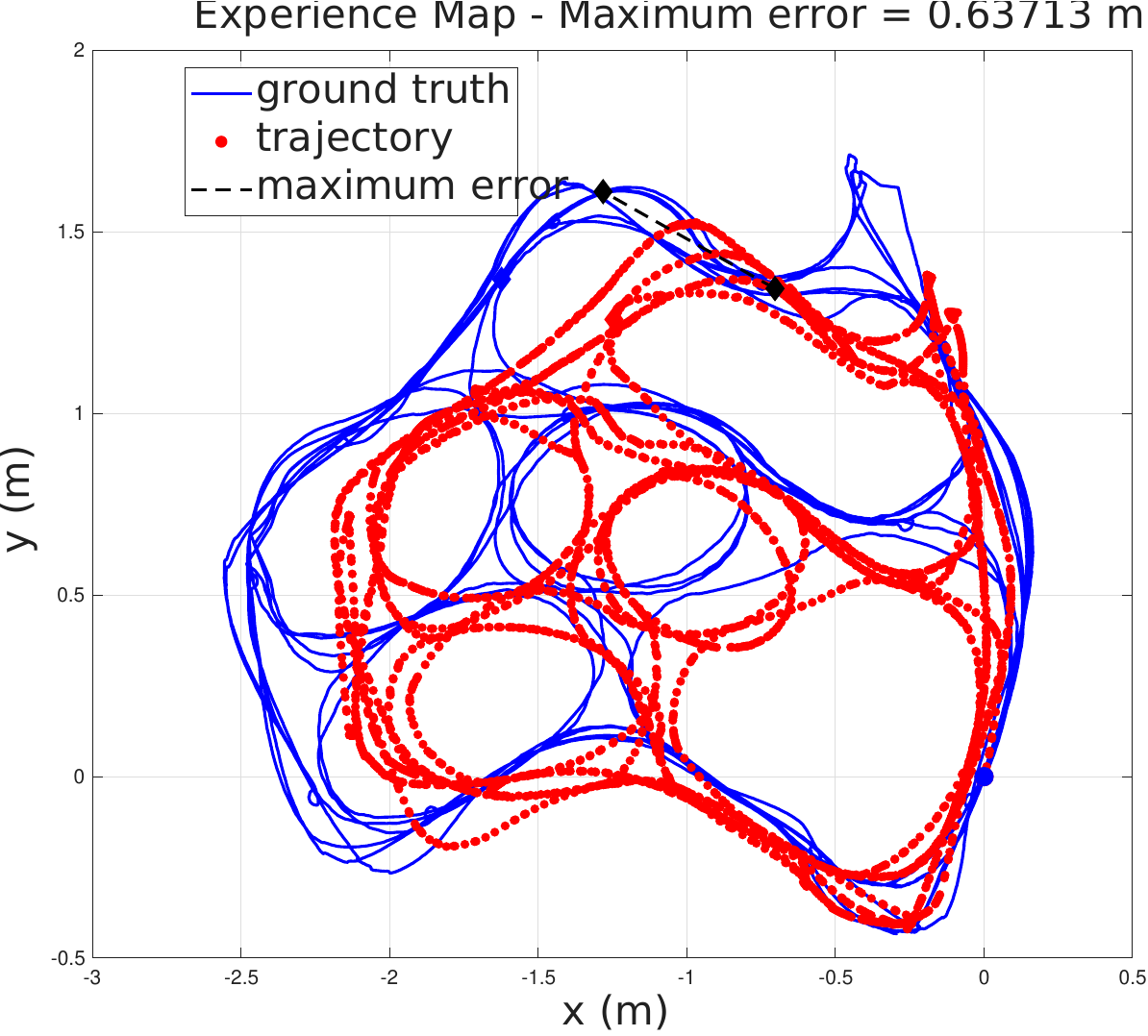}
    \end{minipage}
    \caption{Experience map for iRat Australia dataset by RatSLAM (a) and NeoSLAM (b).}
    \label{fig_exp_map_Irat}
\end{figure}
\autoref{fig_exp_map_Irat}-(a) shows the ground truth obtained from the overhead camera and the maximum error between the two trajectories, in addition to the experience map generated by RatSLAM. \autoref{fig_exp_map_Irat}-(b) presents the result obtained with NeoSLAM. For this dataset, the maximum error of NeoSLAM was slightly higher than that of RatSLAM. Nevertheless, both maps are similar and topologically consistent.


Since we have a reference trajectory for this dataset, the absolute error was calculated for each position between the estimated trajectory and the ground truth, which were temporally synchronized. RatSLAM was slightly superior to NeoSLAM in this dataset according to this metric, with an average error of 0.24 m against 0.29 m for NeoSLAM.

\subsection{FIU MMC Lake Dataset}

\begin{figure}[t]
\centerline{\includegraphics[width=.9\linewidth]{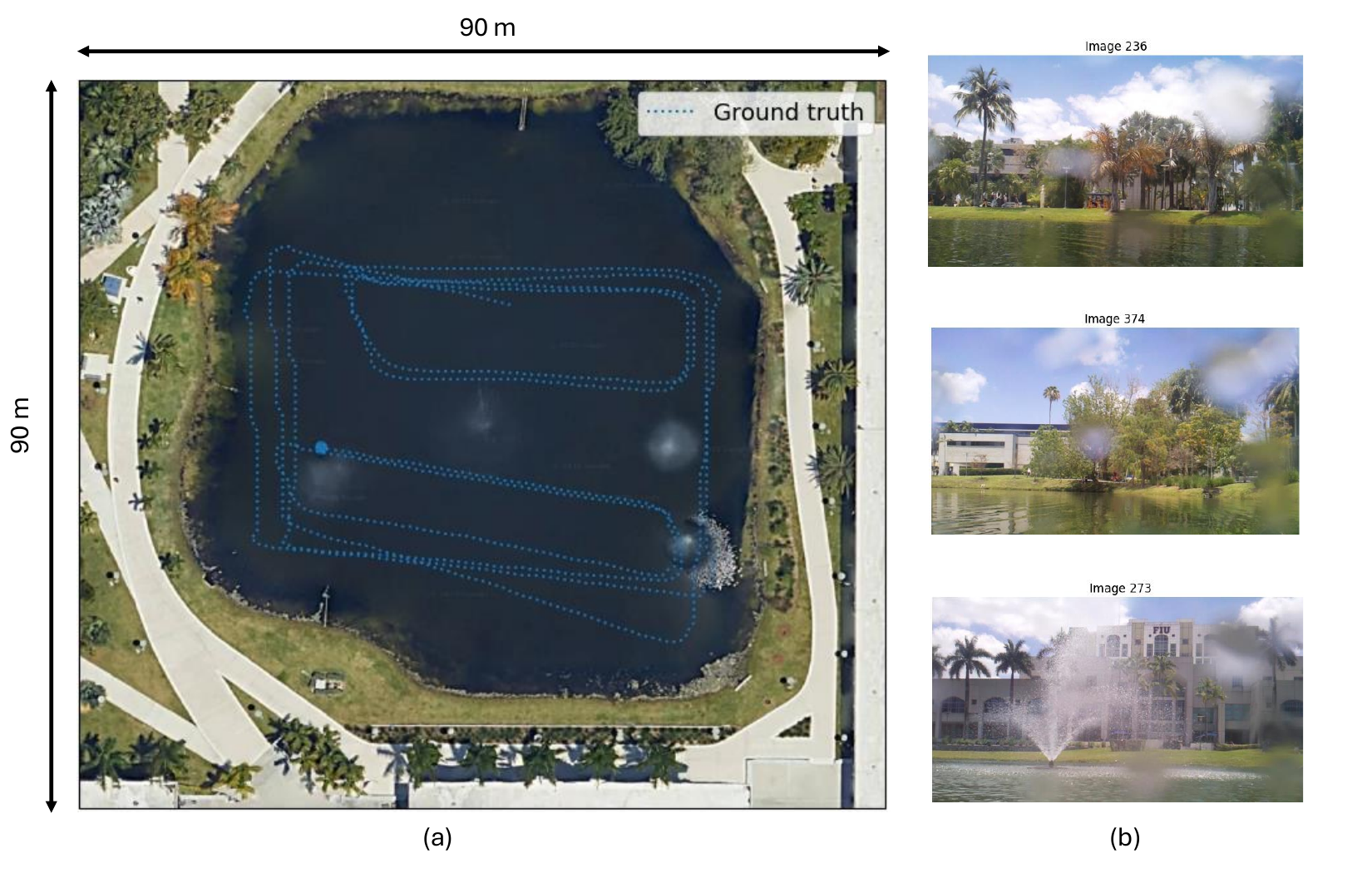}}
\caption{(a) FIU MMC Lake Dataset and (b) frames examples from the frontal camera.}
\label{Lake_Dataset}
\end{figure}

\begin{figure}[t]
    \centering
    \includegraphics[width=.45\linewidth]{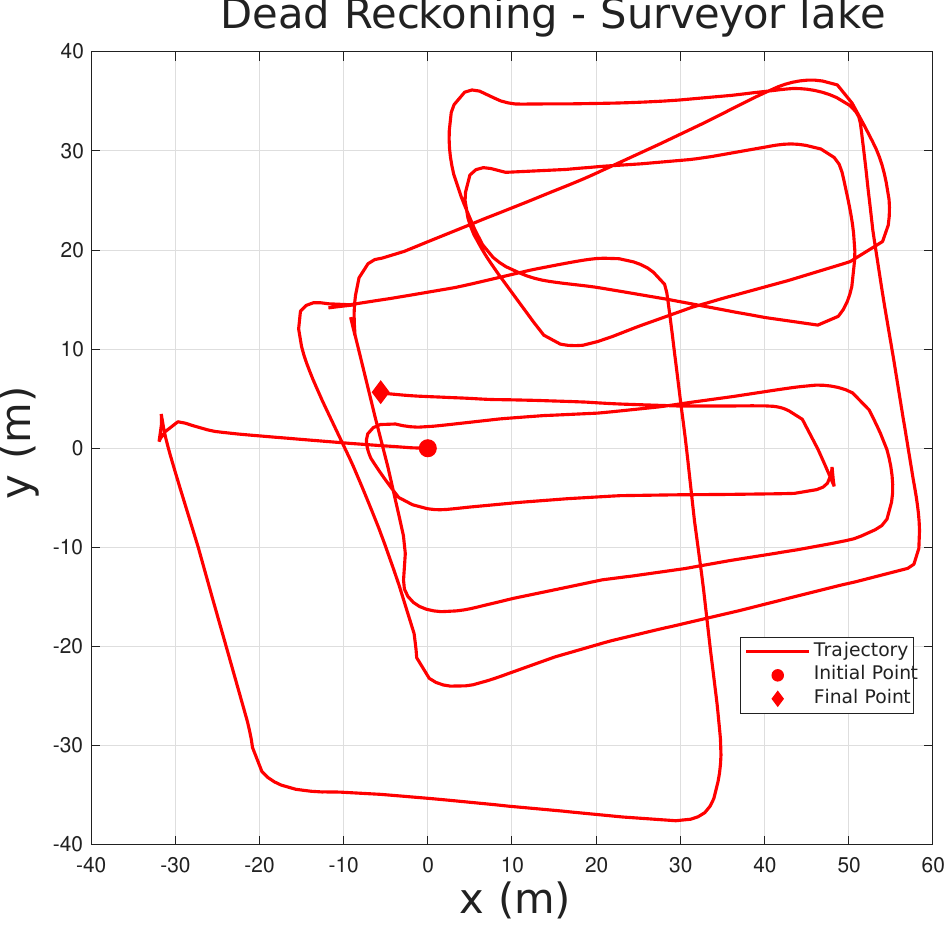}
    \caption{Odometry from Surveyor ASV at Green Library Lake - Florida International University.}
    \label{fig_DeadReckoning_Lake}
\end{figure}

The acquisition of this dataset was detailed in \cite{icar2025}. Data were collected using a SeaRobotics Surveyor USV at the Green Library Lake at Florida International University (FIU), Modesto Maidique Campus (MMC). Figure~\ref{Lake_Dataset} shows the trajectory followed by the USV, which covers approximately 900 meters over approximately 16 minutes, along with some sample images from the frontal camera. The dead reckoning can be observed in figure \ref{fig_DeadReckoning_Lake}.

\begin{figure}[t]
    \begin{minipage}[t]{0.48\linewidth}
        \centering
        \includegraphics[width=1\linewidth]{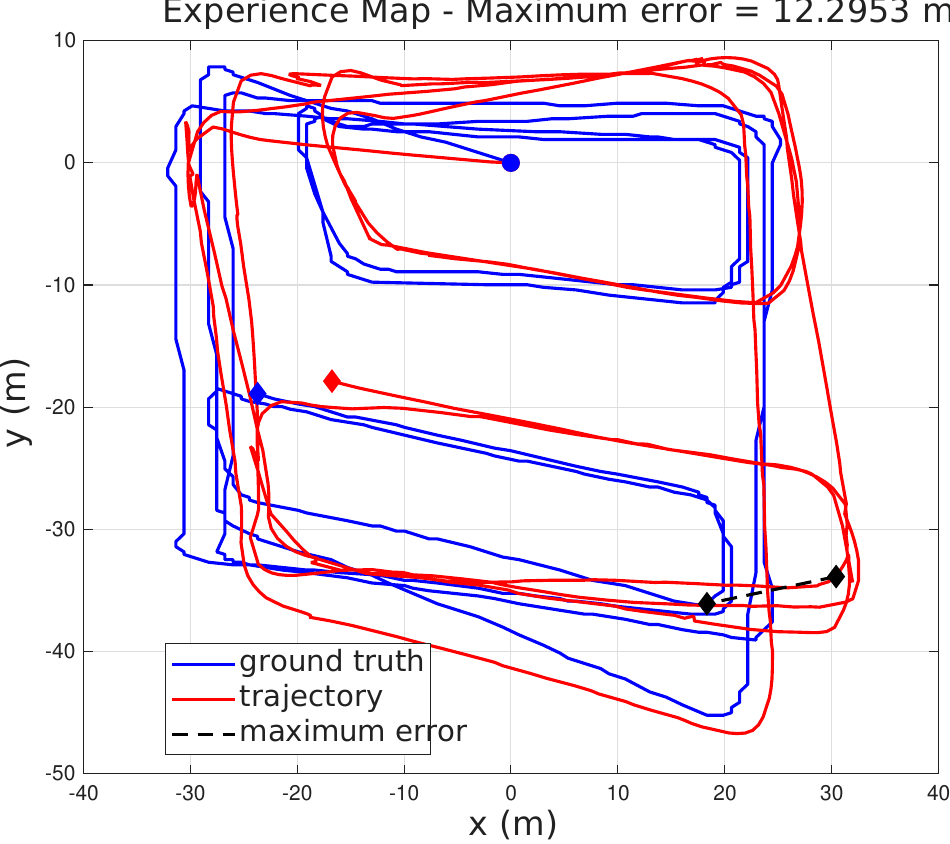}
    \end{minipage}
    \begin{minipage}[t]{0.48\linewidth}
        \centering
        \includegraphics[width=1\linewidth]{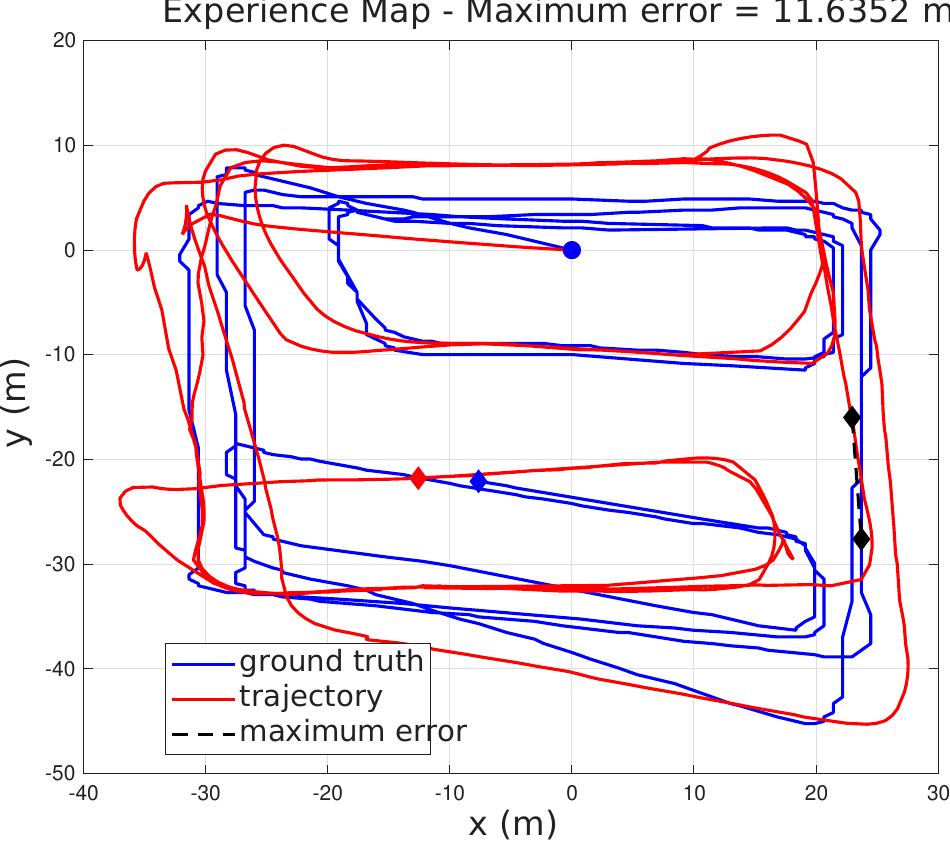}
    \end{minipage}
    \caption{Experience map of the FIU MMC Lake dataset by RatSLAM (a) and NeoSLAM (b).}
    \label{fig_exp_map_Surveyor_lake}
\end{figure}

\autoref{fig_exp_map_Surveyor_lake} shows the trajectories estimated by both algorithms. On the left, we used the same parameters as in \cite{icar2025}, and the result was consistent, represented by the red line. However, the maximum error between the experience map and the ground truth is also shown. It can be observed that the result obtained with NeoSLAM was slightly superior to that of RatSLAM. The average error of NeoSLAM was also lower than that of RatSLAM: 5.92 versus 6.10, respectively.  To achieve this outcome, it was necessary to increase the parameter \verb|exclude_recent_intervals|; otherwise, the robot incorrectly relocalized with nearby earlier parts of the trajectory. 

\section{Conclusion}

This paper presented a new implementation of NeoSLAM, consisting of a complete rewrite into a modular architecture based on modern frameworks. This redesign improves computational efficiency and substantially reduces real-time data loss, allowing the system to close more loops and thereby improving its robustness, while also facilitating maintainability and integration with current tools. The proposed implementation was evaluated against the original NeoSLAM in terms of real-time performance, and against RatSLAM in terms of map reconstruction, across three datasets under varying environmental conditions. To the best of our knowledge, this is the first work evaluating NeoSLAM on a USV dataset, which is particularly challenging due to the purely inertial odometry source and the significant visual ambiguity. Even under these conditions, both NeoSLAM and RatSLAM were able to construct topological maps that improve upon the degraded pure odometry trajectory, and, considering the reconstructed Cartesian trajectories, both methods achieved comparable performance across the evaluated datasets. 

\section*{Acknowledgment}
This research was supported in part by Coordenação de Aperfeiçoamento de Pessoal de Nível Superior - Brasil (CAPES) - Finance Code 001, Conselho Nacional de Desenvolvimento Científico e Tecnológico (CNPq), Fundação Carlos Chagas Filho (FAPERJ) under grant E-26/204.669/2024, and NSF grants IIS-2024733 and IIS-2331908, the Office of Naval Research grant N00014-23-1-2789, the U.S. Department of Defense grant 78170-RT-REP, and the Florida Department of Environmental Protection grant INV31.

\end{document}